\documentclass[11pt]{article}

\usepackage[preprint]{acl}

\usepackage{times}
\usepackage{latexsym}
\usepackage[T1]{fontenc}
\usepackage[utf8]{inputenc}
\usepackage{microtype}
\usepackage{inconsolata}
\usepackage{graphicx}
\usepackage{booktabs}
\usepackage{amssymb}
\usepackage{array}
\usepackage{tabularx}
\usepackage{placeins}
\usepackage{float}
\newcolumntype{Y}{>{\raggedright\arraybackslash}X}

\title{Cognitive Extensions for Dual-Process Language Agents:\\ Memory and Self-Reflection in Interactive Environments}
\author{
\textbf{João Meneses dos Santos\textsuperscript{1}},
\textbf{Arlindo L. Oliveira\textsuperscript{1,2}}
\\
\\
\textsuperscript{1}Instituto Superior Técnico, Universidade de Lisboa, Portugal
\\
\\
\textsuperscript{2}INESC-ID, Lisboa, Portugal
}

\begin{document}
\maketitle

\begin{abstract}
Language agents remain brittle in interactive environments, where success requires long-horizon state tracking, valid action execution, and recovery from failed steps. We extend SwiftSage, a dual-process agent that combines a fast action proposer with a slower planner, using two modular cognitive extensions: an Adaptive Memory Module (AMM) for salience-gated episodic storage and trigger-driven retrieval, and a Self-Reflection Module (SRM) for bounded execution-time validation and corrective intervention. Both modules are implemented as feature-flagged extensions over the same execution substrate, enabling controlled ablations on ScienceWorld. Across four configurations---baseline, baseline+AMM, baseline+SRM, and the full system---the full system achieves the best mean final score (64.62), success rate (43.17\%), and successful-step efficiency (19.33 steps), while SRM is the strongest standalone contributor. The results suggest that execution-time control is the dominant bottleneck in this setting, while episodic memory becomes most useful once the runtime loop is stabilized.
\end{abstract}

\section{Introduction}
Large language models can generate fluent text and solve many short-horizon reasoning problems, but these abilities do not automatically yield reliable agentic behavior. Interactive environments require an agent to maintain state over time, choose actions that are valid in the current world, decompose goals into executable subgoals, recover from unexpected observations, and avoid locally plausible but unproductive loops. The central problem is therefore not only what an agent knows, but how it coordinates fast action proposal, slower deliberation, memory, and execution-time monitoring. A common failure pattern in such environments is that a reasonable high-level plan degrades at execution time: the agent repeats a stale action, proposes an action unavailable in the current state, or continues navigating despite no measurable progress. These failures are difficult to solve with larger prompts alone because they occur at the interface between language generation and environment transition.

This work studies this problem through a dual-process language-agent architecture. SwiftSage~\cite{lin2023swiftsagegenerativeagentfast} provides a natural baseline because it separates \emph{Swift}, a fast System~1-style action proposer, from \emph{Sage}, a slower System~2-style planner whose outputs are executed through an action buffer. This design improves efficiency over methods that query a large model at every timestep, but it leaves two important limitations in long-horizon tasks. First, the agent lacks persistent episodic memory: it cannot selectively reuse salient prior experience across episodes. Second, it does not systematically validate actions immediately before execution or intervene in a bounded way when behavior stagnates.

We address these limitations with two modular extensions. The \emph{Adaptive Memory Module} (AMM) adds salience-gated episodic writing and trigger-driven retrieval. It records compact episodes after informative transitions, such as success, positive progress, near misses, or invalid-action failures, and retrieves relevant memories only at selected recovery and planning points. The \emph{Self-Reflection Module} (SRM) adds bounded execution-time control. It validates actions immediately before they reach the environment, monitors post-step trajectory signals for stagnation, and invokes a constrained Critic only when corrective intervention is justified.

The modules occupy different causal interfaces. AMM is informational: it changes what the controller can remember and reuse, but it never chooses the final action. SRM is control-oriented: it changes what is allowed to reach the environment and when corrective actions are inserted. This separation makes the resulting ablation study interpretable as cooperation between evidence and control rather than as an opaque redesign of the baseline agent.

We evaluate the system on ScienceWorld~\cite{wang2022scienceworldagentsmarter5th}, an interactive text benchmark requiring agents to perform elementary science tasks through grounded sequential actions. The evaluation compares four configurations under a shared runtime substrate: baseline, baseline+AMM, baseline+SRM, and the full system. The results show that the full system achieves the best aggregate performance, while SRM is the strongest standalone extension. AMM alone yields smaller gains, but its contribution is more coherent once SRM stabilizes execution. Overall, the findings support a precise claim: in this setting, execution-time control is the dominant lever for improving interactive language agents, while episodic memory is most useful when inserted as bounded evidence into an already controlled runtime loop.

\section{Motivation and Related Work}
Dual-process theory distinguishes fast, automatic cognition from slower deliberative reasoning~\cite{Kahneman}. In language-agent design, this distinction is useful as an engineering abstraction rather than a claim of cognitive equivalence: fast pathways support cheap local action proposal, while slower pathways support planning, verification, and recovery. Chain-of-thought prompting and related methods make deliberation more explicit in static reasoning tasks~\cite{wei2023chainofthoughtpromptingelicitsreasoning,press2023measuringnarrowingcompositionalitygap,khot2023decomposedpromptingmodularapproach,zhou2023leasttomostpromptingenablescomplex,wang2023selfconsistencyimproveschainthought,zhang2022automaticchainthoughtprompting}, but interactive agents additionally need to decide when to deliberate, how to ground plans in current state, and how to prevent invalid actions from consuming environment steps.

A second relevant line of work augments LLMs with actions, tools, and feedback, including affordance-grounded action selection in SayCan~\cite{ahn2022icanisay}. ReAct interleaves reasoning traces with environment-facing actions~\cite{yao2023reactsynergizingreasoningacting}, Reflexion uses verbal feedback from prior attempts~\cite{shinn2024reflexion}, Self-Refine iteratively revises model outputs~\cite{madaan2023selfrefineiterativerefinementselffeedback}, and CRITIC verifies and corrects outputs using external tools~\cite{gou2024criticlargelanguagemodels}. Toolformer and ART further show that tool calls can be learned or orchestrated in multi-step reasoning pipelines~\cite{schick2023toolformerlanguagemodelsteach,paranjape2023artautomaticmultistepreasoning}. These approaches show that feedback can improve LLM behavior, but they also motivate boundedness: unconstrained self-correction can increase cost or degrade performance when feedback is unreliable~\cite{huang2023large}. SRM follows this lesson by making reflection trigger-gated and execution-facing rather than always on.

AMM is motivated by Complementary Learning Systems, where rapid episodic learning and slower consolidation play distinct roles~\cite{Marr1971SimpleMA,clshipponeo}, as well as by memory-augmented generation and cognitive-agent memory systems. Retrieval-augmented generation conditions model outputs on external information~\cite{lewis2020retrieval}, while MemGPT treats the context window as a scarce resource and explicitly moves information between transient context and persistent storage~\cite{packer2023memgpt}. CoALA similarly frames language agents in terms of memory, actions, and decision procedures~\cite{sumers2024cognitive}, and long-horizon language-agent systems such as Generative Agents show how memory, reflection, and planning can support coherent behavior over extended interactions~\cite{park2023generativeagentsinteractivesimulacra}. AMM adapts these ideas to ScienceWorld by storing compact episodic records and retrieving them only under operational triggers, instead of using memory as continuous prompt expansion or as an alternative planner.

The closest architectural predecessor is SwiftSage~\cite{lin2023swiftsagegenerativeagentfast}. It already instantiates fast and slow thinking in ScienceWorld: Swift proposes local actions using an efficient model, while Sage performs higher-level planning and grounding through an action buffer. ScienceWorld is especially suitable for this comparison because it evaluates whether science knowledge can be transformed into valid procedures, not merely whether a model can state the right answer. Our contribution is to preserve the SwiftSage substrate while adding two missing mechanisms: persistent episodic reuse and just-in-time execution control. This framing also separates our work from approaches that simply add more reasoning calls. We ask whether memory and reflection improve an already dual-process controller when inserted at bounded, causally interpretable interfaces.

\section{Method}
The proposed system preserves the SwiftSage control loop and inserts AMM and SRM at precise runtime interfaces. At each timestep, the controller executes the next buffered action when available; otherwise it queries Swift, escalating to Sage under baseline conditions such as invalid actions, no-progress behavior, or the need for deliberate planning. AMM can augment selected Swift, Sage, and Critic prompts with retrieved memories. SRM validates actions before execution and can inject bounded corrective actions into the same buffer used by Sage.

\begin{figure*}[t]
    \centering
    \includegraphics[width=1\textwidth,height=1\textheight,keepaspectratio]{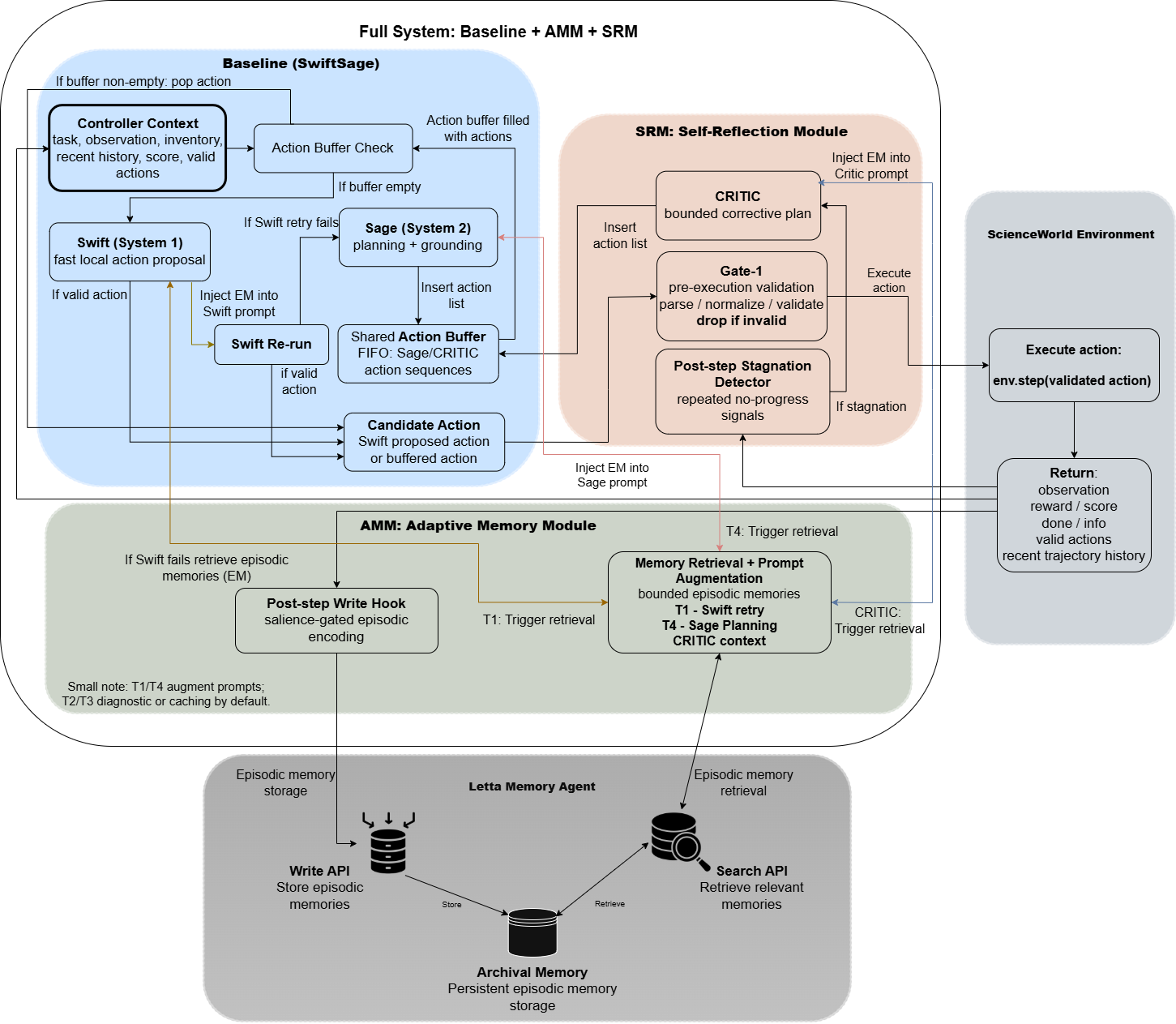}
    \caption{Full-system architecture. SwiftSage remains the action-selection substrate; AMM adds salience-gated episodic writing and trigger-driven retrieval; SRM adds Gate--1 validation, stagnation detection, and bounded Critic correction before actions reach ScienceWorld.}
    \label{fig:full-system-architecture}
\end{figure*}

\subsection{Adaptive Memory Module}
AMM addresses the absence of persistent experience reuse. It has two hook families: a post-step write hook and pre-decision retrieval hooks. The write hook runs after the environment returns an observation and score transition. It receives the task, previous state, executed action, resulting observation, score change, and recent history, then builds a candidate episodic record. This record is stored only if a salience gate detects an informative transition, such as terminal success, positive score change, near-miss progress, explicit invalid-action feedback, or an avoidance-worthy failure.

Stored memories are compact semi-structured records, not raw transcripts. Each memory includes fields such as task, local state, recent context, executed action, resulting observation, score transition, and a type tag. The tag distinguishes success, near-miss, and avoidance-oriented memories. This representation improves retrieval targeting and makes prompt injection safer, because retrieved evidence is already concise, typed, and easy to filter or truncate.

Retrieval is trigger-driven. T1 corresponds to Swift failure: when Swift fails to produce a valid action, AMM retrieves related episodes and retries Swift with memory-conditioned context. T4 corresponds to System~2 planning: when the baseline invokes Sage and memory planning is enabled, AMM retrieves success and near-miss episodes for deliberative planning. T2 and T3 are implemented as stagnation and repeated-invalid-action retrieval/caching hooks, but in the evaluated AMM-only configuration they do not directly modify model inputs. This conservative design avoids injecting weakly grounded negative evidence into the fast pathway.

Prompt augmentation is bounded and explicitly delimited. Swift receives only a small memory block in recovery mode, while Sage may receive a slightly larger block because it is already the deliberative pathway. Prompts instruct the model to treat memories as hints rather than authority: current observations, inventory, admissible actions, and runtime constraints always dominate. Retrieved memories are deduplicated, filtered by operational type, and truncated using a fixed compression order before injection. If retrieval, formatting, or prompt-structure checks fail, execution falls back to the unmodified baseline prompt. AMM therefore changes the evidence available to the controller without changing the final execution channel.

\subsection{Self-Reflection Module}
SRM targets execution-time failures: actions may be plausible in language but invalid, stale, redundant, or ineffective in the current environment. It consists of Gate--1 validation, post-step stagnation detection, and bounded Critic intervention.

Gate--1 is applied immediately before an action reaches the environment. It receives the proposed action, current valid-action set, recent state descriptors, and runtime constraints. It normalizes the action, checks admissibility, applies deterministic repair when the mismatch is minor and safe, and drops the action if it remains invalid or violates constraints. Gate--1 is source-agnostic: Swift actions, Sage-buffered actions, and Critic-generated actions all pass through the same pre-execution control surface.

After each executed step, SRM updates deterministic diagnostics over the recent trajectory. These diagnostics track repeated observations, repeated actions with no effect, invalid-action loops, excessive navigation without state diversity, and verb-level loops that do not change the effective state. When these signals cross a threshold, SRM emits a structured stagnation report summarizing the failure pattern, actions to avoid, and relevant runtime constraints.

The Critic is invoked only when stagnation is detected and safeguards permit a call. Its prompt contains the task, current state, recent trajectory, stagnation report, runtime constraints, and, in the full system, optional AMM evidence. The Critic must output a short executable action list. Its output is parsed, filtered, and inserted into the same FIFO buffer used by Sage; it is never executed directly. Buffered Critic actions still pass through Gate--1. Calls are bounded by budgets and cooldowns, making SRM a controlled repair mechanism rather than an always-on second planner. This is important because the Critic is powerful but risky: if used too often, it can interrupt otherwise valid trajectories or replace local action selection with unnecessary deliberation. SRM therefore treats the Critic as a last-mile repair mechanism, while Gate--1 provides the always-on safeguard.

\subsection{Full-System Composition}
The full system composes AMM and SRM without changing the shared execution substrate. AMM writes salient memories after environment steps and retrieves bounded evidence under its trigger regimes. SRM applies Gate--1 before execution and monitors stagnation after execution. Their main interaction occurs during reflective correction: when SRM invokes the Critic, AMM may provide retrieved episodic evidence as supporting context. The hierarchy remains explicit: current valid actions and constraints dominate; stagnation diagnosis identifies the local failure mode; retrieved memories provide optional evidence; and Gate--1 remains the final safeguard. This design enables a clean comparison between baseline, baseline+AMM, baseline+SRM, and full system. It also makes the composition interpretable: AMM should help when the agent needs relevant prior evidence, especially in recovery and planning regimes; SRM should help when the main failure is execution validity or local stagnation. The full system tests whether these mechanisms are complementary or redundant.

\section{Experimental Setup}
We evaluate on ScienceWorld~\cite{wang2022scienceworldagentsmarter5th}, using tasks 0--29 and up to 10 natural-language variations per task, for up to 271 evaluation episodes per configuration. Each episode runs until completion or a fixed step budget. All configurations use the same environment interface, valid-action grounding, one \texttt{env.step} per timestep, Swift/Sage arbitration, and action-buffer execution. When SRM is enabled, actions additionally pass through Gate--1 and stagnation-triggered Critic actions may be inserted into the buffer. When AMM is enabled, memory writing and trigger-driven retrieval are active.

AMM-enabled configurations are evaluated with a populated memory agent, because the module is designed to reuse prior episodes. The earlier weaker-memory protocol is treated as a sensitivity condition rather than the main architecture-level comparison.

\paragraph{Protocol note.}
We do not claim a new ScienceWorld SOTA; our focus is a controlled ablation over a reproduced SwiftSage-style substrate under a shared local-model/runtime setting. Differences from Lin et al.~\cite{lin2023swiftsagegenerativeagentfast} include the local Qwen2.5-7B-Instruct-1M Sage/Critic runtime and the populated-memory protocol used for AMM-enabled configurations.

We report three primary metrics: final score at termination, task success, and \emph{Steps@Succ}, the mean number of steps conditioned on successful completion. To avoid overweighting tasks with more variations, scores are first aggregated at the task level and then macro-averaged. We also group tasks by oracle trajectory length: Short ($0<*\!Len\leq20$), Medium ($20<*\!Len\leq50$), and Long ($*\!Len>50$). Mechanism-level logs track AMM writes and injections, SRM drops and Critic calls, and controller-usage proxies such as Swift-executed steps, buffer-executed steps, and System~2 call rates. These logs are essential because final scores alone cannot distinguish whether improvement comes from better memory-conditioned planning, fewer invalid actions, shorter successful trajectories, or reduced reliance on expensive System~2 calls.

\begin{table}[t]
\centering
\small
\setlength{\tabcolsep}{3.5pt}
\caption{Aggregate performance across configurations. Final is task-level macro-averaged final score; Success is macro-averaged task success; Steps@Succ is computed only over successful episodes.}
\label{tab:main-summary}
\begin{tabular}{lccc}
\toprule
\textbf{Configuration} & \textbf{Final} & \textbf{Succ.} & \textbf{Steps} \\
\midrule
Baseline       & 51.67 & 23.99\% & 24.48 \\
+AMM           & 53.83 & 26.94\% & 24.03 \\
+SRM           & 64.33 & 41.33\% & 19.76 \\
Full           & \textbf{64.62} & \textbf{43.17\%} & \textbf{19.33} \\
\bottomrule
\end{tabular}
\end{table}

\begin{table*}[t]
\centering
\small
\setlength{\tabcolsep}{5pt}
\caption{Grouped ScienceWorld results. Values are mean final score. Parentheses show relative change over the baseline within each group.}
\label{tab:grouped-results}
\begin{tabular}{lccccc}
\toprule
\textbf{Group} & \textbf{$*\!Len$} & \textbf{Baseline} & \textbf{+AMM} & \textbf{+SRM} & \textbf{Full} \\
\midrule
Short  & 11.76 & 57.74 & 66.43 (+15.1\%) & \textbf{81.50} (+41.2\%) & 80.06 (+38.7\%) \\
Medium & 28.58 & 45.20 & 47.16 (+4.3\%)  & 48.42 (+7.1\%)  & \textbf{49.72} (+10.0\%) \\
Long   & 94.30 & 50.93 & 47.78 (-6.6\%)  & 61.64 (+21.0\%) & \textbf{61.69} (+21.1\%) \\
\midrule
Overall & 49.26 & 51.67 & 53.83 (+4.1\%) & 64.33 (+24.5\%) & \textbf{64.62} (+25.1\%) \\
\bottomrule
\end{tabular}
\end{table*}

\section{Results and Analysis}
Table~\ref{tab:main-summary} shows that the full system obtains the best aggregate performance, with a mean final score of 64.62, success rate of 43.17\%, and Steps@Succ of 19.33. Baseline+SRM is close behind in score, reaching 64.33, while baseline+AMM reaches 53.83 compared with 51.67 for the baseline. Relative to the baseline, the full system improves mean final score by 25.1\%, SRM alone by 24.5\%, and AMM alone by 4.1\%. The central pattern is therefore that SRM accounts for most of the standalone improvement, while AMM provides a smaller positive aggregate effect.

The grouped results in Table~\ref{tab:grouped-results} refine this picture. On Short tasks, baseline+SRM performs best, suggesting that just-in-time execution filtering is especially valuable when trajectories are short and mistakes consume a large fraction of the budget. On Medium and Long tasks, the full system obtains the highest grouped score, although on Long tasks it is nearly tied with SRM. This indicates that memory is most useful in the harder regimes where prior episodes can support recovery or planning, but only after execution has been stabilized.

Success and efficiency show the same trend. Success increases from 23.99\% in the baseline to 26.94\% with AMM, 41.33\% with SRM, and 43.17\% in the full system. Successful trajectories also become shorter in SRM-enabled settings: Steps@Succ falls from 24.48 in the baseline to 19.76 with SRM and 19.33 in the full system. Thus, the strongest configurations do not merely collect more partial reward; they complete more episodes and require fewer steps when they succeed.

Mechanism-level logs explain why. AMM is active in memory-enabled configurations, writing 4.48 memories per episode in AMM-only and 4.89 in the full system. Most writes are near-miss records, meaning the memory store primarily captures partial progress rather than only terminal successes. However, memory is used less frequently once SRM is enabled: T1 triggers fall from 17.13 to 12.15, T4 triggers from 19.13 to 11.77, Swift injections from 30.81 to 23.03, and Sage injections from 19.13 to 11.77. The full system therefore writes slightly more experience but needs fewer memory-conditioned recovery attempts, consistent with SRM reducing the number of unstable trajectories.

SRM shows the complementary profile. SRM-only records 9.77 stagnation reports and 1.35 effective Critic calls per episode; the full system records 9.35 and 1.29. Gate--1 drops are much more frequent: 20.03 per episode in SRM-only and 17.48 in the full system, mostly because proposed actions are not in the current valid-action set. This implies that SRM's main contribution is not frequent reflective replanning, but continuous filtering of invalid or stale actions before they consume environment steps.

Controller usage further supports this interpretation. Direct Swift-executed steps increase from 44.08\% in the baseline to 49.16\% with AMM, 57.13\% with SRM, and 62.01\% in the full system. Buffer-executed steps fall from 55.92\% to 37.99\%, and System~2 call rates fall from 24.6 per episode in the baseline to 14.1 in the full system. Better performance therefore does not come from more deliberation. It comes from making execution more reliable, so fewer invalid actions reach the environment and fewer costly System~2 interventions are required.

Sensitivity analyses reinforce this conclusion. Evaluating AMM with populated memory improves score from 51.47 in a weaker-memory regime to 53.83. Increasing the SRM Critic budget from three to six calls worsens performance, reducing score from 64.33 to 52.95. Disabling Swift-level T1 memory injection in the full system lowers score from 64.62 to 63.33, with the strongest negative effect on Long tasks. These results suggest that memory is useful but should remain targeted, and that reflection must be bounded rather than expanded indiscriminately.

\section{Discussion}
The results support three claims about modular dual-process agents. First, execution-time control is the dominant bottleneck in this ScienceWorld setting. Baseline+SRM nearly matches the full system in final score and produces most of the success-rate and efficiency gains. This is consistent with SRM's position in the runtime loop: it acts immediately before environment interaction, where invalid or stale actions would otherwise consume steps. In contrast, AMM changes the information available to the model but cannot itself ensure that a proposed action is admissible or useful under the current state.

Second, memory is useful but conditional. AMM alone improves the aggregate score only modestly, despite frequent writing and retrieval. This does not imply that episodic memory is irrelevant; rather, it suggests that memory is not the first bottleneck when execution remains unstable. The full system writes slightly more memories but uses memory-conditioned recovery and planning less often than AMM-only. This pattern suggests that SRM reduces the number of unstable states requiring memory intervention, allowing retrieved episodes to be used more selectively.

Third, more reflection is not automatically better. The Critic-budget sensitivity analysis shows that expanding reflective intervention can degrade performance. This supports the design choice of bounded reflection: Gate--1 should remain the cheap always-on safeguard, while the Critic should be reserved for trajectories that exhibit explicit stagnation. The controller-usage statistics point in the same direction. Stronger configurations use fewer System~2 calls and fewer buffer-executed steps, not more. They win by making the fast path safer and the slow path more targeted.

These findings refine the cognitive motivation behind the architecture. The contribution is not that memory and reflection should be added everywhere, but that each mechanism should be placed at a causal interface that matches its role. AMM supplies evidence for recovery and planning. SRM governs execution and corrective intervention. SwiftSage remains the shared fast/slow substrate. This separation is what makes the ablation interpretable and what prevents the full system from becoming a monolithic prompt expansion.

\section{Conclusion}
We introduced two modular cognitive extensions for a SwiftSage-style dual-process language agent: AMM for salience-gated episodic reuse and SRM for bounded execution-time control. Evaluated on ScienceWorld, the full system achieves the best aggregate score, success rate, and successful-step efficiency, while SRM is the strongest standalone contributor. AMM alone is beneficial but modest; its clearest role appears in combination with SRM, where memory can support recovery and planning after execution has been stabilized.

The broader implication is that cognitively inspired agent extensions are most useful when inserted at the right causal interface. Adding memory is not sufficient if the agent still executes invalid or stale actions, and adding reflection is harmful when it is unbounded. In this setting, robust interactive behavior emerges from a disciplined composition: memory provides evidence, reflection controls execution, and the baseline fast/slow controller remains the shared substrate for interpretable ablation.

\section{Limitations}
A first limitation concerns the memory subsystem. AMM currently relies on compact semi-structured episodic records stored in an external memory agent and consumed through bounded prompt injection. This design is appropriate for controlled integration and inspection, but it likely constrains the standalone value of memory: stored traces are useful as situated evidence, yet their representation, retrieval ranking, and conversion into concrete action improvements remain relatively simple. Future work should study richer episodic formats, stronger retrieval and reranking, consolidation across related traces, and more targeted prompt-conditioning strategies for Swift, Sage, and the Critic.

A related limitation is that AMM remains primarily episodic. The system does not yet implement a mature semantic consolidation layer that abstracts across episodes into reusable skills, procedures, or task-general regularities. From a Complementary Learning Systems perspective, the current implementation covers the rapid episodic side more than the slower semantic side. This helps explain why AMM alone is beneficial but modest: the memory store can recall concrete past experiences, but it does not yet reliably transform repeated experiences into generalized procedural knowledge.

A second limitation concerns SRM. The control layer is intentionally interpretable and rule-bounded, but several choices remain hand-crafted, including stagnation thresholds, action-filtering rules, focus-related constraints, fixed Critic budgets, and deterministic repair policies. Gate--1 is effective as a conservative safeguard, but its repairs are shallow when an invalid action cannot be cleanly normalized or mapped to an admissible alternative. Learned verifiers, confidence-aware escalation, richer affordance models, and adaptive reflective budgets may improve this component while preserving the boundedness that proved important in the current experiments.

A third limitation is evaluation scope. The experiments are restricted to ScienceWorld, a text-based simulated science environment, and to the implementation choices inherited from the underlying SwiftSage-style runtime. Although ScienceWorld is well aligned with the target failure modes, the results may not transfer directly to other interactive environments, multimodal settings, robotic action spaces, or different model backbones. In addition, AMM-enabled configurations are evaluated with a populated memory agent, which is appropriate for testing experience reuse but does not fully characterize cold-start learning, warm-up dynamics, memory aging, compaction, or forgetting.

Finally, the analysis remains partly correlational at the prompt level. Mechanism-level logs show when memories are written, retrieved, and injected, and when SRM drops actions or invokes the Critic, but they do not fully isolate every causal prompt-level factor behind individual successes or failures. More fine-grained causal tracing, counterfactual replay, and controlled prompt ablations would strengthen the interpretability of future evaluations.

\section{Ethical Considerations}
This work studies agents in a simulated educational science environment and does not involve human subjects or private user data. The main risks are indirect: techniques that improve autonomous action selection, recovery from failure, and long-horizon persistence could be transferred to less controlled settings. For that reason, the proposed modules emphasize bounded intervention, explicit action validation, logging, and environment-grounded constraints rather than unconstrained autonomous planning. The system should not be interpreted as safe for deployment in open-ended real-world environments without additional oversight, safety evaluation, and domain-specific constraints.

\bibliographystyle{acl_natbib}
\bibliography{custom}
\clearpage
\onecolumn
\appendix
\setlength{\textfloatsep}{0.8\baselineskip}
\setlength{\intextsep}{0.7\baselineskip}
\setlength{\abovecaptionskip}{0.45\baselineskip}
\setlength{\belowcaptionskip}{0.25\baselineskip}

\section{Reproducibility Notes}
\label{app:reproducibility}
The main paper reports the architecture-level findings. This appendix provides implementation details, complete tables, and sensitivity checks needed to interpret the reported results. The evaluation uses ScienceWorld tasks 0--29 with up to ten natural-language variations per task, yielding up to 271 evaluation episodes per configuration. All configurations share the same ScienceWorld interface, valid-action grounding, one-\texttt{env.step}-per-timestep execution semantics, Swift/Sage arbitration, and FIFO action-buffer execution. Reported metrics are computed over evaluation episodes only and are aggregated task-first before macro-averaging across groups.

\paragraph{Memory initialization and leakage control.}
AMM is evaluated as an experience-reuse mechanism rather than as a cold-start learner. The protocol initializes the memory agent with a warm-up stage of two variations per task, i.e., 60 episodes in total; these warm-up episodes are used only to populate memory and are excluded from all reported evaluation metrics. The main AMM comparison uses a populated memory agent, matching the operating condition of the full system, while the weaker-memory regime is reported separately in Table~\ref{tab:app-sensitivity}. This distinction is central to interpreting AMM: the results measure retrieval from an initialized episodic store, not learning from an empty memory. During an episode, the online score is determined by ScienceWorld before any post-step AMM write can affect later behavior; post-step writes may contribute to later retrieval but do not retroactively affect the score of the transition that generated them.

\paragraph{Runtime constants and model settings.}
The executable substrate keeps Swift as the fast action proposer and Sage as the slow planner. Swift follows the SwiftSage-style Flan-T5-large behavior-cloning component, while Sage and the SRM Critic use Qwen2.5-7B-Instruct-1M served locally through vLLM. AMM stores salient episodes in Letta archival memory. The active memory-consumption caps are three retrieved episodes for Swift T1 recovery and five retrieved episodes for Sage T4 planning; Critic outputs are parsed as short corrective action lists of up to five actions and are inserted into the same FIFO buffer as Sage plans. The main SRM setting uses a maximum Critic budget of three calls per episode; the six-call setting is reported only as a sensitivity variant in Table~\ref{tab:app-sensitivity}.

\paragraph{Aggregation and table organization.}
Tables~\ref{tab:app-full-results} and~\ref{tab:app-success-efficiency} are task-level macro-averages: variations are averaged within each task, and task values are then averaged within oracle-length groups. Tables~\ref{tab:app-amm-activity} and~\ref{tab:app-srm-activity} are task-balanced per-episode mechanism summaries. Table~\ref{tab:app-usage} reports configuration-level usage proxies, with System~2 calls corresponding to Sage calls in Baseline/AMM and to Sage+Critic calls in SRM/Full. The appendix is organized in the same order as the main analysis: full scores, success and efficiency, mechanism activity, usage proxies, and sensitivity analyses. Each block includes a short interpretation paragraph immediately after the corresponding table.

\section{Score Results by Task}
\label{app:full-results}
Table~\ref{tab:app-full-results} expands the grouped result table from the main paper with the full per-task score breakdown.

\begin{table}[H]
\centering
\scriptsize
\setlength{\tabcolsep}{4pt}
\caption{Full ScienceWorld score table. Values are mean final score over variations. $\ast$Len is the average oracle trajectory length used to define Short, Medium, and Long groups. Parentheses in grouped rows show relative change over the baseline within each group.}
\label{tab:app-full-results}
\begin{tabular}{@{}lrrrrr@{}}
\toprule
\textbf{Task (Type)} & \textbf{$\ast$Len} & \textbf{Baseline} & \textbf{+AMM} & \textbf{+SRM} & \textbf{Full} \\
\midrule
1-1 (L) & 107.7 & 71.56 & 55.78 & 71.67 & 61.89 \\
1-2 (L) & 78.6 & 42.22 & 61.22 & 58.67 & 47.67 \\
1-3 (L) & 88.9 & 38.44 & 38.44 & 50.33 & 46.00 \\
1-4 (L) & 75.2 & 51.22 & 69.44 & 68.89 & 66.00 \\
2-1 (M) & 21.4 & 84.70 & 75.80 & 74.30 & 72.70 \\
2-2 (M) & 35.2 & 39.80 & 39.70 & 40.00 & 40.20 \\
2-3 (L) & 65.0 & 54.60 & 56.10 & 84.00 & 84.20 \\
3-1 (S) & 13.6 & 48.40 & 64.40 & 71.40 & 75.40 \\
3-2 (M) & 20.8 & 36.40 & 47.00 & 49.00 & 36.80 \\
3-3 (M) & 25.6 & 68.00 & 70.70 & 67.50 & 70.50 \\
3-4 (M) & 29.0 & 64.80 & 70.40 & 73.60 & 70.30 \\
4-1 (S) & 14.6 & 60.80 & 75.00 & 100.00 & 98.30 \\
4-2 (S) & 8.8 & 92.50 & 100.00 & 100.00 & 100.00 \\
4-3 (S) & 12.6 & 45.80 & 59.10 & 92.50 & 90.80 \\
4-4 (S) & 14.6 & 75.80 & 90.00 & 100.00 & 100.00 \\
5-1 (L) & 69.5 & 26.10 & 23.90 & 10.80 & 11.30 \\
5-2 (L) & 79.6 & 24.60 & 23.40 & 70.60 & 85.40 \\
6-1 (M) & 33.6 & 29.75 & 35.75 & 30.50 & 37.00 \\
6-2 (S) & 15.1 & 27.78 & 27.78 & 30.00 & 37.78 \\
6-3 (M) & 23.0 & 11.33 & 10.33 & 10.67 & 28.44 \\
7-1 (S) & 7.0 & 85.00 & 80.00 & 95.00 & 95.00 \\
7-2 (S) & 7.0 & 55.00 & 70.00 & 100.00 & 100.00 \\
7-3 (S) & 8.0 & 63.10 & 74.80 & 93.30 & 93.30 \\
8-1 (M) & 40.0 & 26.80 & 27.60 & 41.80 & 41.80 \\
8-2 (S) & 16.3 & 23.25 & 23.25 & 32.75 & 10.00 \\
9-1 (L) & 97.0 & 59.00 & 54.00 & 66.00 & 68.00 \\
9-2 (L) & 84.9 & 62.00 & 51.50 & 55.00 & 59.00 \\
9-3 (L) & 123.1 & 73.00 & 55.50 & 62.00 & 69.00 \\
10-1 (L) & 130.1 & 41.70 & 50.60 & 66.70 & 75.10 \\
10-2 (L) & 132.1 & 66.70 & 33.50 & 75.00 & 66.70 \\
\midrule
Short (S) & 11.76 & 57.74 & 66.43 (+15.1\%) & 81.50 (+41.2\%) & 80.06 (+38.7\%) \\
Medium (M) & 28.58 & 45.20 & 47.16 (+4.3\%) & 48.42 (+7.1\%) & 49.72 (+10.0\%) \\
Long (L) & 94.30 & 50.93 & 47.78 (-6.6\%) & 61.64 (+21.0\%) & 61.69 (+21.1\%) \\
Overall & 49.26 & 51.67 & 53.83 (+4.1\%) & 64.33 (+24.5\%) & 64.62 (+25.1\%) \\
\bottomrule
\end{tabular}
\end{table}

\vspace{-0.4em}
\noindent\footnotesize\emph{Reading note.} Improvements are not uniformly distributed. AMM is competitive on selected tasks, but the largest and most consistent gains appear in SRM-enabled runs. The full system is strongest overall because it preserves SRM's execution-time control while allowing memory to support recovery and planning on longer trajectories.

\normalsize\medskip

\section{Success and Successful-Step Efficiency}
\label{app:success-efficiency}
Table~\ref{tab:app-success-efficiency} reports completion and efficiency metrics using the same oracle-length grouping as the main score analysis.

\begin{table}[H]
\centering
\scriptsize
\setlength{\tabcolsep}{4pt}
\caption{Task success and efficiency, macro-averaged within oracle-length groups. Success is the fraction of successful episodes; Steps@Succ is the mean number of steps conditioned on successful completion.}
\label{tab:app-success-efficiency}
\begin{tabular}{@{}lrrrrrrrr@{}}
\toprule
& \multicolumn{2}{c}{\textbf{Baseline}} & \multicolumn{2}{c}{\textbf{+AMM}} & \multicolumn{2}{c}{\textbf{+SRM}} & \multicolumn{2}{c}{\textbf{Full}} \\
\cmidrule(lr){2-3}\cmidrule(lr){4-5}\cmidrule(lr){6-7}\cmidrule(l){8-9}
\textbf{Group} & \textbf{Succ.\%} & \textbf{Steps} & \textbf{Succ.\%} & \textbf{Steps} & \textbf{Succ.\%} & \textbf{Steps} & \textbf{Succ.\%} & \textbf{Steps} \\
\midrule
Short (S)  & 32.96 & 11.77 & 46.59 & 10.78 & 64.77 & 8.09  & 73.90 & 9.75 \\
Medium (M) & 11.94 & 20.75 & 11.94 & 44.13 & 11.94 & 31.00 & 8.96  & 26.67 \\
Long (L)   & 25.86 & 37.37 & 20.69 & 39.96 & 40.52 & 29.70 & 39.70 & 31.89 \\
Overall    & 23.99 & 24.48 & 26.94 & 24.03 & 41.33 & 19.76 & 43.17 & 19.33 \\
\bottomrule
\end{tabular}
\end{table}

\vspace{-0.4em}
\noindent\footnotesize\emph{Reading note.} The full system has the best overall completion rate and the best overall Steps@Succ. Short tasks show the clearest completion gain, while Long tasks show that SRM-enabled systems complete substantially more episodes than the baseline and require fewer steps when they succeed. Medium tasks are less stable: the full system improves final score but not success rate.

\normalsize\medskip

\section{Mechanism-Level Activity}
\label{app:mechanisms}
Tables~\ref{tab:app-amm-activity} and~\ref{tab:app-srm-activity} report the mechanism-level activity that supports the interpretation of the aggregate results.

\begin{table}[H]
\centering
\scriptsize
\setlength{\tabcolsep}{4pt}
\caption{AMM activity by oracle-length group. Values are per-episode task-balanced averages for memory writes, retrieval triggers, and Swift/Sage memory injections.}
\label{tab:app-amm-activity}
\begin{tabular}{@{}llrrrrrrrr@{}}
\toprule
\textbf{Config.} & \textbf{Group} & \textbf{Writes} & \textbf{Success} & \textbf{Near-miss} & \textbf{Avoid.} & \textbf{T1} & \textbf{T4} & \textbf{Swift inj.} & \textbf{Sage inj.} \\
\midrule
AMM  & Short   & 3.63 & 1.38 & 1.79 & 0.47 & 18.43 & 20.12 & 33.62 & 20.12 \\
AMM  & Medium  & 4.48 & 0.86 & 2.87 & 0.75 & 21.58 & 26.05 & 40.64 & 26.05 \\
AMM  & Long    & 5.18 & 0.65 & 3.37 & 1.16 & 13.09 & 13.69 & 21.91 & 13.69 \\
AMM  & Overall & 4.48 & 0.95 & 2.71 & 0.82 & 17.13 & 19.13 & 30.81 & 19.13 \\
\midrule
Full & Short   & 3.96 & 1.96 & 1.92 & 0.37 & 10.91 & 10.33 & 21.00 & 10.33 \\
Full & Medium  & 5.55 & 0.90 & 3.22 & 1.44 & 19.04 & 16.74 & 34.64 & 16.74 \\
Full & Long    & 5.23 & 0.85 & 3.74 & 0.67 & 8.60  & 9.41  & 16.99 & 9.41 \\
Full & Overall & 4.89 & 1.14 & 2.99 & 0.78 & 12.15 & 11.77 & 23.03 & 11.77 \\
\bottomrule
\end{tabular}
\end{table}

\vspace{-0.4em}
\noindent\footnotesize\emph{Reading note.} AMM writes are dominated by near-miss memories in both memory-enabled configurations, so the store primarily captures partial progress rather than only terminal successes. Once SRM is enabled, AMM writes slightly more but retrieves and injects less often, suggesting that execution control reduces the number of unstable states requiring memory-conditioned recovery.

\normalsize\medskip

\begin{table}[H]
\centering
\scriptsize
\setlength{\tabcolsep}{4pt}
\caption{SRM activity by oracle-length group. Values are per-episode task-balanced averages for stagnation reports, effective Critic calls, main Gate--1 drop reasons, and total Gate DROP counts. Reason-code columns are non-exclusive and therefore do not sum to total drops.}
\label{tab:app-srm-activity}
\begin{tabular}{@{}llrrrrrr@{}}
\toprule
\textbf{Config.} & \textbf{Group} & \textbf{Stagn.} & \textbf{Critic} & \textbf{Not valid} & \textbf{NOOP} & \textbf{Target unseen} & \textbf{Gate DROP} \\
\midrule
SRM  & Short   & 10.47 & 0.67 & 7.45  & 1.02 & 0.00 & 13.75 \\
SRM  & Medium  & 13.11 & 2.20 & 25.48 & 1.38 & 3.37 & 27.00 \\
SRM  & Long    & 6.96  & 1.36 & 20.74 & 2.08 & 1.56 & 25.06 \\
SRM  & Overall & 9.77  & 1.35 & 17.57 & 1.54 & 1.52 & 20.03 \\
\midrule
Full & Short   & 8.61  & 0.72 & 7.50  & 1.30 & 0.10 & 8.80 \\
Full & Medium  & 14.40 & 1.76 & 22.89 & 1.49 & 1.63 & 24.49 \\
Full & Long    & 6.61  & 1.45 & 18.44 & 3.23 & 1.23 & 20.05 \\
Full & Overall & 9.35  & 1.29 & 15.98 & 2.12 & 0.97 & 17.48 \\
\bottomrule
\end{tabular}
\end{table}

\vspace{-0.4em}
\noindent\footnotesize\emph{Reading note.} The dominant SRM mechanism is Gate--1 filtering rather than frequent Critic replanning. Effective Critic calls remain close to one per episode overall, while Gate drops are much larger and are mostly caused by actions absent from the current valid-action set. This supports the claim that execution-time validation is the main source of SRM's improvement.

\normalsize\medskip

\section{System Usage and Cost Proxies}
\label{app:usage}
Table~\ref{tab:app-usage} summarizes controller-usage proxies that complement the score, success, and mechanism-level analyses.

\begin{table}[H]
\centering
\small
\caption{System usage and cost proxies by configuration. System~2 means Sage calls in Baseline/AMM and Sage+Critic calls in SRM/Full.}
\label{tab:app-usage}
\begin{tabular}{@{}lrrrr@{}}
\toprule
\textbf{Config.} & \textbf{Swift \%} & \textbf{Buffer \%} & \textbf{S1 calls} & \textbf{S2 calls} \\
\midrule
Baseline & 44.08 & 55.92 & 40.8 & 24.6 \\
AMM      & 49.16 & 50.84 & 60.9 & 19.1 \\
SRM      & 57.13 & 42.87 & 29.4 & 15.7 \\
Full     & 62.01 & 37.99 & 54.4 & 14.1 \\
\bottomrule
\end{tabular}
\end{table}

\vspace{-0.4em}
\noindent\footnotesize\emph{Reading note.} The best-performing systems do not improve by increasing deliberation. System~2 calls decrease from the baseline to the full system, while the share of direct Swift-executed steps increases. This is consistent with the interpretation that SRM makes the fast path safer and the slow path more selective.

\normalsize\medskip

\section{Sensitivity Analyses}
\label{app:sensitivity}
Table~\ref{tab:app-sensitivity} reports targeted sensitivity analyses. These rows are not additional primary architectures; they test whether the conclusions depend on memory initialization, Critic budget, or Swift-level T1 memory injection.

\begin{table}[H]
\centering
\scriptsize
\setlength{\tabcolsep}{4pt}
\caption{Sensitivity analysis of selected implementation choices. Results are grouped final scores.}
\label{tab:app-sensitivity}
\begin{tabular}{@{}lrrrrp{0.28\textwidth}@{}}
\toprule
\textbf{Variant} & \textbf{Overall} & \textbf{Short} & \textbf{Medium} & \textbf{Long} & \textbf{Role} \\
\midrule
AMM weaker-memory regime & 51.47 & 60.45 & 45.69 & 47.85 & Sensitivity to memory initialization. \\
AMM populated memory & 53.83 & 66.43 & 47.16 & 47.78 & Main AMM comparator. \\
SRM Critic max = 6 & 52.95 & 62.70 & 42.10 & 52.05 & Sensitivity to reflective-intervention budget. \\
SRM Critic max = 3 & 64.33 & 81.50 & 48.42 & 61.64 & Main SRM comparator. \\
Full without Swift T1 memory injection & 63.33 & 79.23 & 53.88 & 56.36 & Sensitivity to Swift-level AMM recovery. \\
Full system & 64.62 & 80.06 & 49.72 & 61.69 & Main full-system comparator. \\
\bottomrule
\end{tabular}
\end{table}

\vspace{-0.4em}
\noindent\footnotesize\emph{Reading note.} The Critic-budget comparison is the strongest warning against unbounded reflection: increasing the maximum number of Critic calls from three to six lowers performance in every oracle-length group. The memory-initialization comparison shows that AMM benefits from a more mature episodic store, while the T1 ablation indicates that Swift-level memory recovery contributes most clearly on Long tasks.

\normalsize\medskip

\end{document}